\documentclass{bmvc2k}

\usepackage{booktabs}
\usepackage{multirow}
\usepackage{amsmath}
\usepackage{amssymb}
\usepackage{tikz}
\usepackage{bm} 
\usepackage{diagbox}

\title{Stain-Aware Wavelet Regularization for Instant Adversarial Purification in Histopathology}

\addauthor{Zhe Li}{zhe.li@fau.de}{1}
\addauthor{Bernhard Kainz}{bernhard.kainz@fau.de}{1,2}

\addinstitution{
FAU Erlangen-Nürnberg\\
 Erlangen, Germany
}
\addinstitution{
  Imperial College London\\
  London, UK
 }

\runninghead{Z Li, B Kainz}{SAWR}

\def\etal{\emph{et al}\bmvaOneDot}

\begin{document}

\maketitle

\begin{abstract}
Deep learning has become prevalent in computational pathology pipelines that support tasks such as cancer screening and digital pathology analysis. However, the susceptibility of neural networks to adversarial perturbations raises safety concerns for reliable deployment in clinical practice.
In histopathological images, this challenge is exacerbated by the difficulty of distinguishing high-frequency adversarial noise from subtle and diagnostically relevant tissue structures.
To address this issue, we propose Stain-Aware Wavelet Regularization (SAWR), an adversarial purification framework that leverages multi-level wavelet-domain regularization based on Haar transform to hierarchically disentangle adversarial perturbations from diagnostic structural information. This spectral constraint is further extended to individual histological channels, enabling stain-specific frequency regulation consistent with the biological properties of Hematoxylin and Eosin. Extensive experiments demonstrate that SAWR improves adversarial robustness by up to 10.69\% over the baseline approach, while maintaining texture and spectral fidelity under adversarial perturbations.

\end{abstract}

\section{Introduction}
\label{sec:intro}

Deep learning has been widely adopted in computational pathology, enabling automated systems to achieve strong performance across a range of histopathological tasks, such as cancer subtype classification, tissue segmentation, and grading of malignancy. These methods have demonstrated remarkable diagnostic accuracy on large curated benchmarks and their integration into clinical decision support systems has accelerated substantially in recent years. The success of such systems is largely attributable to the representational power of deep neural networks, which are capable of extracting hierarchical features from high-resolution whole-slide images at a scale that exceeds human capacity for exhaustive review.

Despite these advances, a growing body of evidence in both computer vision~\cite{goodfellow2014explaining,moosavi2016deepfool} and medical imaging~\cite{finlayson2019adversarial,ma2021understanding,bortsova2021adversarial,apostolidis2022survey} indicates that deep neural networks remain highly vulnerable to adversarial perturbations. These perturbations consist of small and often imperceptible input modifications that can induce erroneous predictions with high confidence. Critically, the magnitude of such perturbations can be made sufficiently small that they are invisible to the human eye, but their effect on neural network predictions can be catastrophic. 
In the context of medical image analysis, such vulnerabilities pose significant safety concerns since adversarial manipulation or even incidental acquisition noise may lead to incorrect diagnoses with severe clinical consequences. 
Therefore, ensuring robustness to adversarial perturbations is an important consideration for the reliable deployment of learning-based systems in pathology workflows.
The histopathological domain introduces additional challenges that are absent in natural image settings. Whole-slide images are characterized by high spatial resolution and strong staining variability across institutions and tissue preparation protocols, and a rich set of diagnostically relevant microstructures whose visual signatures are encoded in fine-grained texture patterns. 
These nuclear and chromatin-level cues are predominantly manifested at intermediate and high spatial frequencies, which overlap with the frequency range exploited by adversarial perturbations. As a result, pathology images are particularly susceptible to adversarial corruption that degrades diagnostic utility.

\begin{figure}[tb]
\begin{center}
\includegraphics[width=\textwidth]{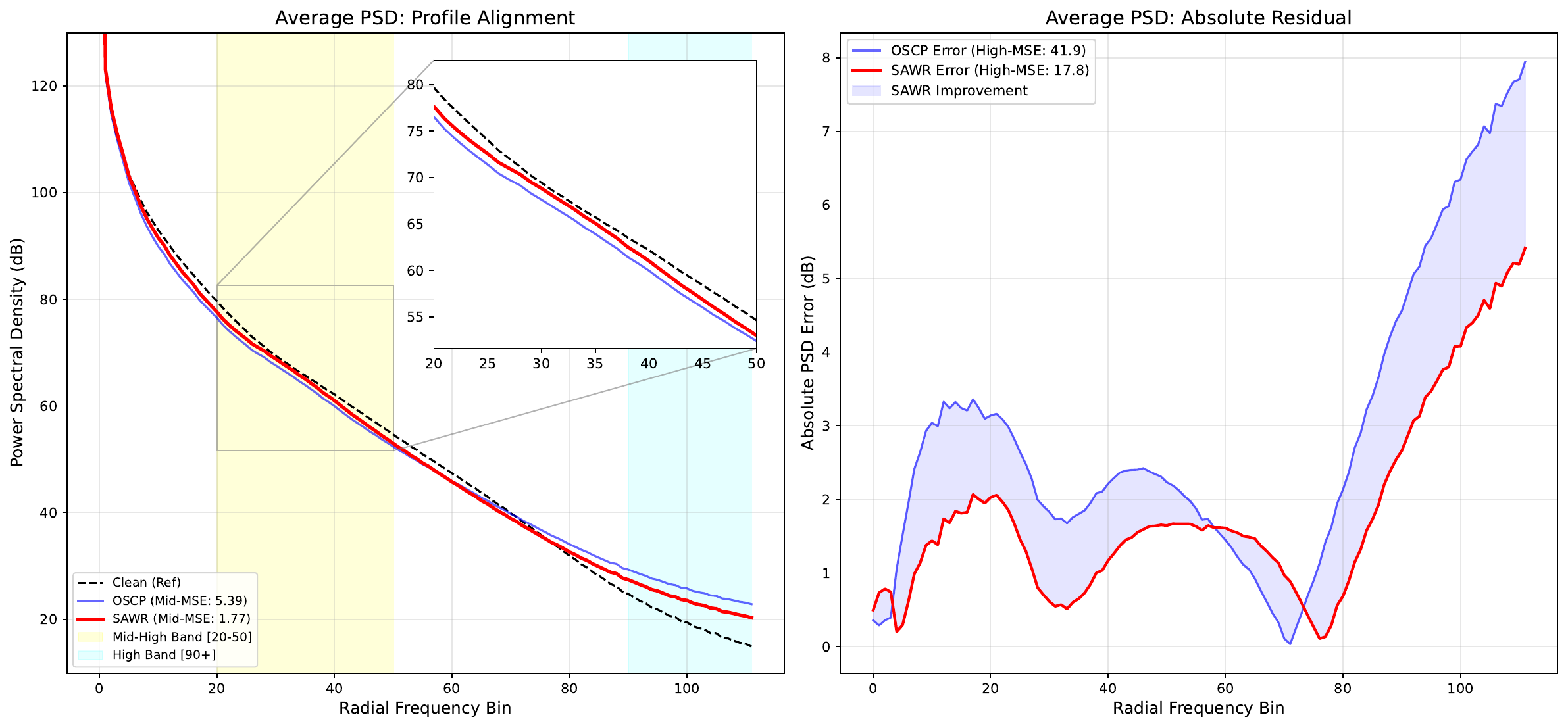}
\end{center}
\caption{Average Power Spectral Density (PSD) analysis over the PathMNIST test set. The profile alignment (left) shows that OSCP deviates from the clean reference in both mid-frequency and high-frequency bands, while SAWR maintains closer spectral alignment across all frequencies. The residual error plot (right) confirms that SAWR achieves substantially lower spectral error than OSCP across the full frequency spectrum (High-Band MSE: 17.8 vs.\ 41.9).}
\label{fig:teaser} 
\end{figure}

Building upon this insight, researchers have explored defense strategies for medical imaging tasks. Ghaffari Laleh~\etal~\cite{ghaffari2022adversarial} investigate adversarial robustness through pathology-aware adversarial training, demonstrating improved robustness while reflecting a trade-off with reduced clean image accuracy. 
In parallel, adversarial purification methods~\cite{nie2022diffusion,wang2022guided} have become predominant in natural image domains by leveraging generative models to remove adversarial perturbations prior to classification. 
Subsequently, OSCP~\cite{lei2025instant} exploits latent consistency models (LCMs~\cite{luo2023latent} and LCM-LoRA~\cite{luo2023lcm}) to substantially reduce computation cost, enabling practical purification under latency constraints.
However, transferring general adversarial purification methods to histopathological images poses unique difficulties due to the domain shift and the reliance on fine-grained texture statistics such as nuclear chromatin patterns and stromal organization. 
The Power Spectral Density (PSD) analysis shown in Figure~\ref{fig:teaser} reveals that the directly transferred method exhibits a noticeable attenuation in mid-frequency bands while retaining excessive energy in ultra-high-frequency regions. This spectral distortion is particularly detrimental in the pathology setting, since mid-frequency components carry essential structural information about tissue architecture and cellular organization that is critical for accurate diagnosis. 

To address these challenges, we propose Stain-Aware Wavelet Regularization (SAWR), a purification framework that incorporates multi-level wavelet-domain regularization into the training objective of a diffusion model. By leveraging hierarchical frequency-band decomposition, SAWR is designed to preserve diagnostically relevant structural details while suppressing adversarial perturbations. 
This frequency-aware design is further adapted to the biochemical characteristics of histopathological images through a stain-aware decomposition module, which applies wavelet constraints to the Hematoxylin and Eosin components separately. Hematoxylin and Eosin stains encode complementary biological information, with Hematoxylin revealing nuclear morphology and chromatin texture while Eosin reflects the cytoplasmic and extracellular matrix organization. Treating these components jointly during regularization risks conflating their distinct frequency profiles and introducing cross-channel artifacts, whereas decoupled regularization allows the framework to impose stain-specific constraints that are aligned with the underlying tissue biology. In addition, a semantic consistency constraint is introduced to encourage the preservation of high-level pathological features during purification, thereby preventing the generative model from producing images that are spectrally plausible but semantically inconsistent with the original tissue context.

Our main contributions are summarized as follows:
\begin{enumerate}
\item We introduce Stain-Aware Wavelet Regularization (SAWR), a biologically motivated adversarial purification framework that integrates multi-level wavelet-domain regularization with a latent consistency-distilled diffusion model.
\item We propose a differentiable stain decomposition method that decouples Hematoxylin and Eosin components, which enables stain-aware regularization in the frequency domain. This decomposition reflects the distinct biochemical and structural information carried by each stain channel.
\item We design a semantic consistency constraint to align feature representations between purified and clean images, thereby preserving pathological semantic information.
\end{enumerate}

\section{Related work}

\subsection{Adversarial attacks and defenses}
Adversarial training (AT) improves model robustness by incorporating adversarially perturbed examples into the training objective~\cite{goodfellow2014explaining,madry2017towards}. Although recent advances have achieved notable gains through large-scale data augmentation and architectural refinements~\cite{gowal2021improving,singh2023revisiting}, AT methods generally exhibit a trade-off between robustness and clean accuracy and tend to be constrained to specific attack types. 
In contrast, adversarial purification offers a defense paradigm that preprocesses inputs to eliminate perturbations before classification, without modifying the classifier itself.
Early approaches based on generative adversarial networks and variational autoencoders~\cite{samangouei2018defense,Schott19} demonstrated the feasibility of this direction but suffered from limited scalability. Diffusion-based purification has since emerged as the dominant approach: DiffPure~\cite{nie2022diffusion} applies a forward diffusion step followed by reverse denoising, while GDMP~\cite{wang2022guided} and related methods introduce guided reverse sampling to improve semantic recovery. Despite strong empirical performance, these methods incur substantial inference latency due to iterative sampling. OSCP~\cite{lei2025instant} reduces this overhead through latent consistency distillation, yet the underlying generative process remains susceptible to oversmoothing and loss of fine-grained structural detail, a limitation that is especially pronounced when evaluated under the comprehensive AutoAttack~\cite{croce2020reliable} benchmark.

\subsection{Adversarial robustness in medical image analysis}

Adversarial robustness has become an increasingly important concern in medical image analysis. Early work~\cite{finlayson2019adversarial} demonstrates that medical machine learning systems are highly susceptible to adversarial attacks, which raises critical safety and reliability issues. Subsequent studies have systematically characterized vulnerability across diverse attack settings and clinical modalities
~\cite{bortsova2021adversarial,dong2024survey,ma2021understanding}. 
Histopathological images present a particularly challenging setting due to the presence of diagnostically informative fine-grained textures, staining variability across acquisition protocols, and the spectral overlap between adversarial perturbations and high-frequency structural cues. PLIP~\cite{10635610} highlights significant adversarial vulnerabilities in multimodal pathology foundation models, while Ghaffari Laleh \etal~\cite{ghaffari2022adversarial} demonstrate that histopathological classifiers are especially susceptible to adversarial corruption and investigate pathology-aware adversarial training as a mitigation strategy. However, these defenses remain subject to the robustness-accuracy trade-off inherent to training-time approaches. Our work instead explores purification as a post-hoc defense strategy and addresses the domain-specific challenges that arise when adapting generative purification to histopathological images. A key component of this adaptation is the explicit modeling of stain structure. Methods such as Macenko~\cite{macenko2009method} and Vahadane~\cite{vahadane2016structure} provide a principled basis for decomposing Hematoxylin and Eosin components and we build upon this foundation to introduce stain-aware regularization within the purification framework.

\subsection{Wavelet-based image analysis}
The Discrete Wavelet Transform provides a joint spatial and frequency representation that has been widely adopted in image analysis tasks requiring multi-scale feature extraction~\cite{mallat2002theory,strang1996wavelets,daubechies1992ten}. Unlike the Fourier transform, which provides only global frequency information, the DWT preserves spatial locality, enabling localized analysis of frequency content at multiple resolution levels. This property makes it particularly well-suited to capturing the hierarchical structure of natural and biomedical images.
In deep learning, wavelet transforms have been integrated into convolutional architectures to replace or augment standard pooling operations~\cite{liu2018multi}, with demonstrated benefits in preserving high-frequency detail and reducing aliasing artifacts. In the medical imaging domain, wavelet-based methods have been applied to tasks including texture analysis, segmentation, and stain normalization, where the ability to disentangle frequency components is diagnostically valuable. Our work builds on this foundation by incorporating multi-level DWT as an explicit regularization objective within a diffusion-based purification framework, rather than as a feature extraction component, and further extends it to operate on stain-decomposed image channels to exploit the domain-specific frequency structure of H\&E histopathology.

\section{Method}

\begin{figure}[tb]
\begin{center}
\includegraphics[width=\textwidth]{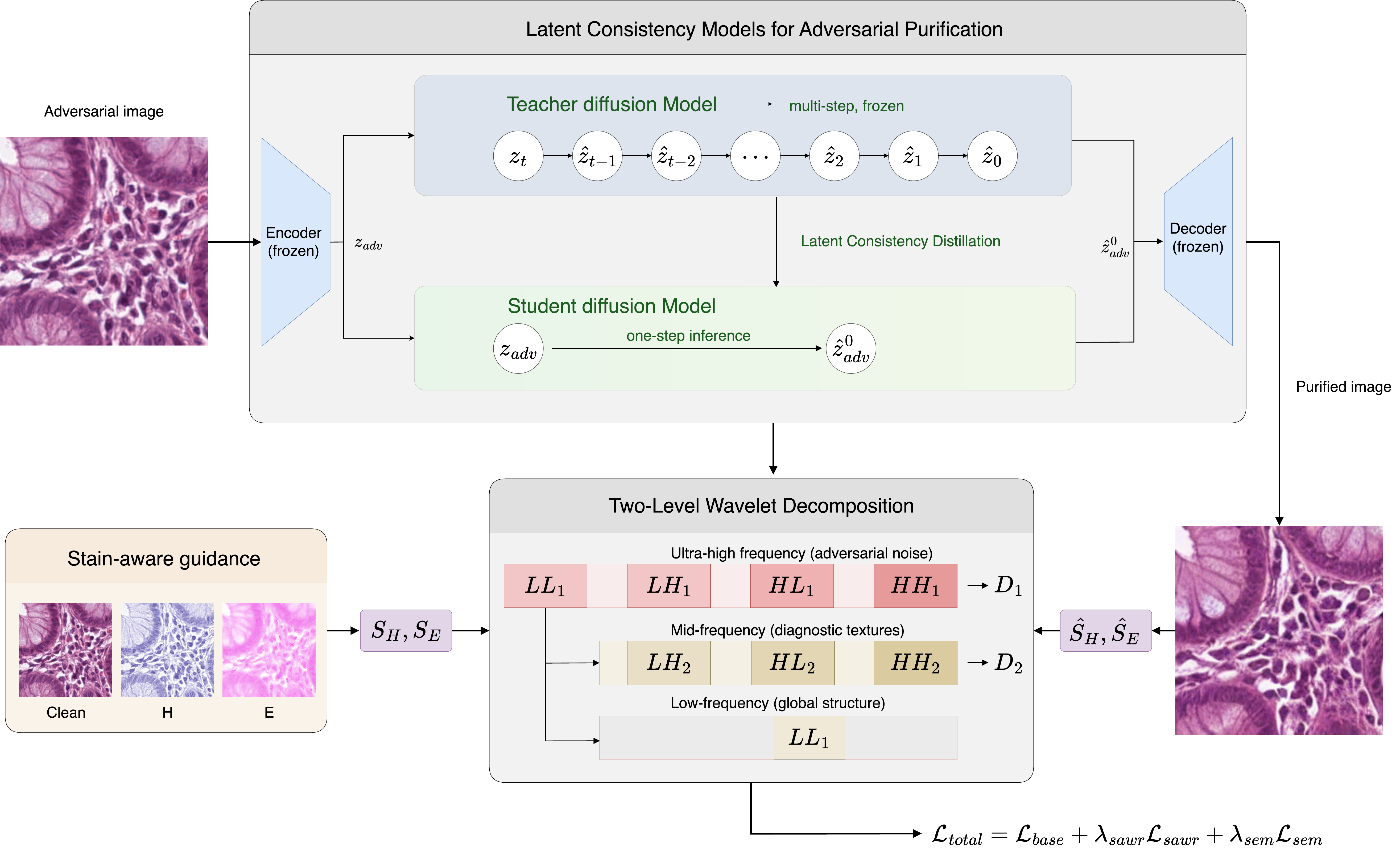}
\end{center}
\caption{Overview of the proposed SAWR framework. During training, the teacher model, encoder, and decoder in the light blue are frozen. The student model purifies the adversarial latent \(z_{adv}\) to \(\hat{z}_{adv}^0\) via latent consistency distillation.  The stain-aware wavelet regularization is integrated into the optimization objectives. The efficacy of this purification is validated through a Power Spectral Density (PSD) curve analysis.} 
\label{fig:stain_framework}
\end{figure}

In this section, we present SAWR, an adversarial purification 
framework for histopathological images illustrated in 
Fig.~\ref{fig:stain_framework}. SAWR builds upon the Instant 
Adversarial Purification framework~\cite{lei2025instant} and 
introduces three domain-specific components to address the 
biochemical and frequency characteristics of H\&E stained tissue. 
The base consistency training objective is described in 
Sec.~\ref{sec:preliminaries}. The stain-aware wavelet regularization is presented in Sec.~\ref{sec:sawr} and the semantic 
feature preservation loss is introduced in Sec.~\ref{sec:sem}.

\subsection{Preliminaries}
\label{sec:preliminaries}
Adversarial purification aims to recover a clean image $x$ from its perturbed counterpart $x_{adv}$ by leveraging the iterative denoising steps of diffusion models. 
To reduce the computation cost, OSCP~\cite{lei2025instant} integrates Latent Consistency Models (LCM) to achieve purification in a single inference step.
The core of OSCP lies in treating adversarial perturbations as a shift in the standard Gaussian noise distribution, modeling the adversarial latent $z_t^*$ as a combination of clean signal, Gaussian noise $\epsilon$, and adversarial noise $\delta_{adv}$ as in Eq.~\ref{eq7:noisecombination}. In inference, 
the latent consistency model maps adversarial latents $z_{adv}$ back to the clean manifold $\hat{z}_{adv}^0$ as in Eq.~\ref{eq7:lcmdenoising}. 

\begin{equation}
\label{eq7:noisecombination}
z_{t}^* = \sqrt{\bar{\alpha}_{t}}z + \sqrt{1-\bar{\alpha}_{t}}(\epsilon + \delta_{adv})
\end{equation}
\begin{equation}
\label{eq7:lcmdenoising}
\hat{z}_{adv}^0 = c_{out}(t) \left( \frac{z_{adv}(t) - \sqrt{1-\bar{\alpha}_t} \hat{\epsilon}_\theta(z_{adv}, c_{ce}, t)}{\sqrt{\bar{\alpha}_t}} \right)
\end{equation}

where $\epsilon \sim \mathcal{N}(0, I)$ and $\hat{\epsilon}_\theta$ is the predicted noise, $\delta_{adv}$ denotes the adversarial perturbation.
The objective of OSCP consists of two components: the Gaussian Adversarial Noise Distillation (GAND) loss \(\mathcal{L}_{\mathrm{GAND}}\) for latent consistency, and the Clear Image Guide (CIG) loss \(\mathcal{L}_{\mathrm{CIG}}\) for stabilizing the purification toward the clean reference $z$.  

\begin{equation}
\label{eq:lossbase}
\mathcal{L}_{\mathrm{base}} = \mathcal{L}_{\mathrm{GAND}} + \lambda_{\mathrm{CIG}} \mathcal{L}_{\mathrm{CIG}}
\end{equation}

Following OSCP, we adopt the combined objective as our base reconstruction loss, denoted as $\mathcal{L}_{\mathrm{base}}$.


\subsection{Stain-aware wavelet regularization}
\label{sec:sawr}
To address the discrepancy between the clean image $x_{clean}$ and purified images $f_\theta(x_{adv})$ in frequency domain, we implement a multi-level Discrete Wavelet Transform (mDWT) using the Haar basis~\cite{strang1996wavelets,daubechies1992ten} to regularize the consistency training of the latent diffusion model $f_\theta$ across distinct frequency bands.
DWT provides a joint spatial and frequency representation that enables localized frequency analysis~\cite{mallat2002theory}. In 2D image processing, it decomposes an image into four sub-bands, $x_{LL}$ (Low-Low), $x_{LH}$ (Low-High), $x_{HL}$ (High-Low), and $x_{HH}$ (High-High), by applying low pass (L) and high pass (H) filters along the two spatial dimensions. The $x_{LL}$ sub-band captures coarse structural information, while $x_{LH}$ and $x_{HL}$ encode directional edge details, and $x_{HH}$ represents fine-scale variations and high frequency noise.  
Wavelet-based architectures~\cite{liu2018multi} demonstrate that integrating DWT into CNNs effectively captures multi-scale features while preserving spatial information.

\paragraph{Multi-level wavelet decoupling.}
In our framework, we employ a recursive decomposition scheme with $J=2$ levels. Each subsequent level operates on the $LL$ subband from the previous level, which yields a coarse-to-fine frequency hierarchy. In the first level ($j=1$), the input image \(x \in \mathbb{R}^{C \times H \times W}\) is decomposed into high-frequency detail subbands $\mathcal{D}_1 = \{LH_1, HL_1, HH_1\}$ and a low-frequency approximation \(LL1 \in \mathbb{R}^{C \times H/2 \times W/2}\). The detail subbands \(\mathcal{D}_1\) capture the highest-frequency components of the signal that correspond to sharp edges and fine textures at the original image resolution. Adversarial perturbations are typically constrained to small \(\ell_\infty\) or \(\ell_2\) norms in pixel space and tend to manifest prominently in these high-frequency subbands. Meanwhile, $LL_1$ preserves coarse structural information and retains the low-frequency envelope of the image content.
In the second level ($j=2$), $LL_1$ is recursively decomposed into mid-frequency details $\mathcal{D}_2 = \{LH_2, HL_2, HH_2\}$ and a deeper approximation \(LL2 \in \mathbb{R}^{C \times H/4 \times W/4}\). At this scale, the detail subbands \(\mathcal{D}_2\) encapsulate mid-range structural patterns in the context of histopathology images. These correspond to fine-grained tissue structures such as chromatin texture patterns, nuclear boundary distributions, and glandular arrangement details that are diagnostically informative and occupy a frequency range distinct from both pixel-level perturbation noise and global layout information. The subband $LL_2$ captures the global tissue layout and encodes region-level spatial organization such as the distribution of cellular density.
Based on this hierarchical frequency stratification, we formulate the multi-level wavelet consistency loss:

\begin{equation}
\label{eq:losswave}
\begin{split}
\mathcal{L}_{\mathrm{mDWT}} = & \sum_{j=1}^{J} \omega_j \sum_{k \in \mathcal{D}_j} || \mathcal{W}_j(f_\theta(x_{\mathrm{adv}}))_k - \mathcal{W}_j(x_{\mathrm{clean}})_k ||_1 \\
& + \omega_{LL} || LL_J(f_\theta(x_{\mathrm{adv}})) - LL_J(x_{\mathrm{clean}}) ||_1
\end{split}
\end{equation}

Here, $\mathcal{W}_j(\cdot)_k$ denotes the $k$-th subband coefficient map at decomposition level $j$, \(||\cdot||_1\) denotes the element-wise $L_1$ distance, and $LL_J(\cdot)$ represents the low-frequency approximation at the deepest level \(J\). The L1 norm is adopted because it is less sensitive to outlier coefficient values arising from localized perturbation artifacts in the wavelet domain. The hyperparameters $\omega_j$ and $\omega_{LL}$ control the relative strength assigned to each frequency band. In detail, $\omega_1$ governs the finest-scale detail subbands where adversarial energy is most concentrated, $\omega_2$ weights the mid-frequency structural subbands, and $\omega_{LL}$ regulates deviation in the residual low-frequency approximation to preserve global image consistency. By jointly supervising across all frequency bands rather than operating solely in the pixel domain, \(\mathcal{L}_{\mathrm{mDWT}}\) encourages the purified output \(f_\theta(x_{\mathrm{adv}})\) to recover spectral fidelity at multiple scales simultaneously.

\paragraph{Stain-specific frequency adaptation.}
In Hematoxylin and Eosin (H\&E) histopathology, diagnostic information is revealed through the differential staining of tissue components, where Hematoxylin (H) highlights the nuclear morphology that forms the diagnostic foreground, while Eosin (E) provides the essential cytoplasmic and stromal context.
Because these two stain channels encode fundamentally different spatial frequency content and carry unequal diagnostic weight, a uniform frequency regularization strategy across the full RGB image is suboptimal. We therefore design channel-specific frequency constraints that apply stronger regularization to the H channel to preserve diagnostic nuclear detail while relaxing constraints on the E channel.

The stain decomposition follows the Beer-Lambert law and is 
implemented as a differentiable operation within the training pipeline. Each image is first converted to optical density space 
via $\mathrm{OD} = -\log(I/255)$. The stain density maps 
$S_H$ and $S_E$ are obtained by applying a fixed stain matrix 
$\mathbf{M} \in \mathbb{R}^{3\times3}$ as 
$\mathbf{C} = \mathrm{OD} \cdot \mathbf{M}^{-1}$, following the 
method of Macenko et al.~\cite{macenko2009method}. 
Since all operations consist of matrix multiplications and element-wise 
transformations, gradients flow through the decomposition into the purification network $f_\theta$ without interruption.
Then, we formulate the stain-aware wavelet regularization (SAWR) loss by extending the general spectral objective $\mathcal{L}_{\mathrm{mDWT}}$ 
(Eq.~\ref{eq:losswave}) independently to each stain component.

For the Hematoxylin channel, we apply the full multi-level wavelet regularization between the target stain map \(S_H(x_{\mathrm{clean}})\) and the predicted map \(\hat{S}_H = S_H(f_\theta(x_{\mathrm{adv}}))\):

\begin{equation}
\label{eq:lossstainH}
\begin{split}
\mathcal{L}_{H} = \mathcal{L}_{\mathrm{mDWT}}(S_H(x_{\mathrm{clean}}), \hat{S}_H; \Omega_H) = &\ \omega_{l1} \sum_{k \in \mathcal{D}_1} || \mathcal{W}(S_H)_k - \mathcal{W}(\hat{S}_H)_k ||_1 \\
& + \omega_{l2} \sum_{k \in \mathcal{D}_2} || \mathcal{W}(S_H)_k - \mathcal{W}(\hat{S}_H)_k ||_1 \\
& + \omega_{ll} || S_{H, LL_2} - \hat{S}_{H, LL_2} ||_1
\end{split}
\end{equation}

where $\Omega_H = \{\omega_{l1}, \omega_{l2}, \omega_{ll}\}$ controls the relative importance of different wavelet subbands. 
For the Eosin channel, the diagnostic relevance of fine-grained high-frequency detail is comparatively limited, as stromal and cytoplasmic structures are characterized by their large-scale spatial distribution rather than sharp local boundaries. Enforcing strict high-frequency consistency on the E channel risks over-constraining the purification network, potentially impeding its ability to suppress adversarial noise in those bands. Therefore, we restrict the regularization in Eosin channel to the low frequency subband \(LL_1\), which encodes the coarse tissue architecture while remaining insensitive to fine-grained perturbation artifacts:

\begin{equation}
\label{eq:lossstainE}
\mathcal{L}_{E} = || S_{E, LL_1} - \hat{S}_{E, LL_1} ||_1
\end{equation}
\begin{equation}
\label{eq:lossstaintotal}
\mathcal{L}_{\mathrm{sawr}} = \lambda_H \mathcal{L}_{H} + \lambda_E \mathcal{L}_{E}
\end{equation}

The SAWR loss is defined as a weighted combination of the two stain-channel objectives as Eq.~\ref{eq:lossstaintotal}, where \(\lambda_H\) and \(\lambda_E\) are hyperparameters that balance the relative frequency regularization.

\subsection{Semantic feature preservation}
\label{sec:sem}
Although \(\mathcal{L}_{\mathrm{base}}\) and \(\mathcal{L}_{\mathrm{sawr}}\) effectively constrain model fidelity and structured signal consistency in the pixel and frequency domains,
discriminative patterns relevant to tissue-type classification remain unaddressed by either objective.
Pixel-level and frequency-level consistency losses operate on low-level signal statistics and do not directly enforce alignment of the high-level semantic content that downstream pathology classifiers rely upon. 
A purified image may closely match the clean image in both pixel intensity and wavelet coefficient distributions while still exhibiting subtle deviations in deep feature space that are sufficient to alter classification decisions.

To address this, we introduce a semantic consistency constraint operating in the latent space of a pre-trained ResNet-50 encoder \(\Phi_{\mathrm{path}}(\cdot)\), which has been trained on histopathology data and encodes domain-relevant semantic representations. Specifically, the constraint aligns the intermediate feature representations of the purified adversarial sample \(f_\theta(x_{\mathrm{adv}})\) and the corresponding clean image \(x_{\mathrm{clean}}\) by minimizing the mean squared error between their feature representations in the latent space:


\begin{equation}
\mathcal{L}_{\mathrm{sem}} = \mathrm{MSE}\bigl(\Phi_{\mathrm{path}}(f_\theta(x_{\mathrm{adv}})),\, \Phi_{\mathrm{path}}(x_{\mathrm{clean}})\bigr)
\end{equation}

The encoder \(\Phi_{\mathrm{path}}(\cdot)\) is kept frozen during training, so its feature space provides a fixed semantic reference that guides purification without being affected by the optimization of \(f_\theta\). 
Our final training objective integrates latent reconstruction, stain-aware wavelet regularization, and feature-level semantic alignment:

\begin{equation}
\mathcal{L}_{\mathrm{total}} = \mathcal{L}_{\mathrm{base}} + \lambda_{\mathrm{sawr}}\,\mathcal{L}_{\mathrm{sawr}} + \lambda_{\mathrm{sem}}\,\mathcal{L}_{\mathrm{sem}}
\end{equation}

where \(\lambda_{\mathrm{sawr}}\) and \(\lambda_{\mathrm{sem}}\) are hyperparameters that control the contribution of the frequency-domain and semantic components, respectively. 
This joint supervision strategy ensures that the purification model \(f_\theta\) restores adversarially corrupted inputs across pixel, frequency, and semantic dimensions simultaneously.

\section{Experiments}

\subsection{Dataset and metrics} 
\paragraph{Dataset.}
We evaluate our framework on the PathMNIST dataset, which is a component of the MedMNIST benchmark~\cite{yang2023medmnist}. PathMNIST is derived from a large-scale colorectal cancer tissue slide collection and contains high-resolution (224 \(\times\) 224) histopathological images categorized into 9 colon tissue types, including adipose tissue, background, debris, lymphocytes, mucus, smooth muscle, normal colon mucosa, cancer-associated stroma, and colorectal adenocarcinoma epithelium~\cite{kather2019predicting}.
The dataset is partitioned into 89,996 training samples and 7,180 test samples with approximately balanced class distribution.

\begin{table}[tb]\setlength{\tabcolsep}{6pt} 
\begin{center}
\begin{tabular}{llccl}
\toprule
Attacks  & Methods & Clean Acc & Robust Acc & Defense \\
\hline
\multirow{3}{*}{untargeted PGD-100}  & --& 94.15 & 0 & Without defense \\
 & OSCP~\cite{lei2025instant} & 81.39 & 70.24 & Adv. Purification \\
 & SAWR (ours) & \textbf{86.10} & \textbf{79.75} & Adv. Purification \\
 \hline
\multirow{3}{*}{targeted PGD-40}  & -- & 94.15 & 0.11 &  Without defense \\
 & OSCP~\cite{lei2025instant} & 81.39 & 78.65 & Adv. Purification \\
 & SAWR (ours) & \textbf{86.10} & \textbf{84.68} & Adv. Purification \\
 \hline
\multirow{3}{*}{AutoAttack}  & -- & 94.15 & 0 & Without defense \\
 & OSCP~\cite{lei2025instant} & 81.39 & 67.08 & Adv. Purification \\
 & SAWR (ours) & \textbf{86.10} & \textbf{77.77} & Adv. Purification \\
\bottomrule
\end{tabular}
\end{center}
\caption{Results on PathMNIST. The classifier is ResNet-50. }
\label{tab:sotastainwave}
\end{table}

\paragraph{Attack setup.} 
To evaluate the robustness of our purification method, we consider adversarial attacks under the $L_{\infty}$ norm constraint. We denote the maximum allowed perturbation per pixel as $\epsilon$ and the step size as $\eta$. Following standard protocols, we utilize both Projected
Gradient Descent (PGD)~\cite{madry2017towards} and the more rigorous AutoAttack~\cite{croce2020reliable} for evaluation.
For the untargeted PGD-100 attack, we set the perturbation budget \(\epsilon = 4/255\) and step size \(\eta = 1/255\) over 100 iterations. In the targeted PGD-40 attack, we apply a larger budget of \(\epsilon = 16/255\) with a step size of \(\eta = 0.4/255\), which simulates a threat model where the adversary has a specific target class in mind. AutoAttack evaluation employs the standard ensemble including APGD-CE and APGD-DLR, FAB, and the Square attack, with a budget of \(\epsilon = 4/255\) and \(\eta = 1/255\). AutoAttack minimizes the need for manual tuning of attack hyperparameters such as step size and thus provides a more reliable indicator of worst-case robustness than single-method attacks.

\paragraph{Implementation details. }
Prior to training the LCM, we fine-tune a ResNet-50 as both the classifier and the white-box attacker for 10 epochs on the clean PathMNIST training set using the Adam optimizer with a learning rate of \(1\times10^{-4}\). Stable Diffusion v1.5~\cite{rombach2022high} is fine-tuned on the clean PathMNIST training set to function as the teacher model for consistency distillation. The LCM is trained with a learning rate of \(8 \times 10^{-6}\) using a constant scheduler with an initial warmup phase of 500 steps.  The hyperparameters are set to \(\lambda_{\mathrm{sawr}} = 0.01\) and \(\lambda_{\mathrm{sem}} = 0.01\). The wavelet subband weights are set to \(\omega_{l1} = 2.0\), \(\omega_{l2} = 2.0\), and \(\omega_{ll} = 1.0\) for the Hematoxylin channel, with \(\lambda_H = 2.0\) and \(\lambda_E = 1.0\) in the SAWR loss. The larger value of $\lambda_H$ reflects the greater diagnostic importance of nuclear morphology in H\&E tissue classification. At inference, SAWR functions as a model-agnostic and plug-and-play purifier that requires no modification to the downstream classifier. We report clean accuracy on the original images and robust accuracy on purified adversarial images to measure classification performance.
All experiments are conducted on a single A100 GPU. 
The baseline OSCP takes about 12 hours for 10,000 steps while our method converges at epoch 2 (step 8,460 with batch size 32) in all experiments and takes about 8 hours.
Despite adding two extra loss terms, training speed did not decrease due to an optimized implementation and fewer steps. No additional inference cost is introduced compared to baseline.

\subsection{Comparison with state-of-the-art.} 
\noindent\textbf{Quantitative comparison.}
Table~\ref{tab:sotastainwave} summarizes the performance of our proposed SAWR compared to the baseline. The ResNet-50 classifier achieves a clean accuracy of 94.15\% when trained on original images. However, this model exhibits extreme vulnerability to adversarial perturbations and drop to near-zero robust accuracy under all attack settings.
This result demonstrates that standard training alone is insufficient to ensure robustness in clinical settings.

Given the absence of adversarial purification frameworks specifically designed for medical imaging, we adopt OSCP~\cite{lei2025instant} as a strong baseline for instant purification originally developed for natural images. We exclude adversarial training methods due to their sacrifice of clean accuracy, and other diffusion-based purification mehtods that require a large number of sampling steps and result in high computational cost.
As shown in Table~\ref{tab:sotastainwave}, SAWR consistently outperforms the baseline across all attack scenarios. 
Under the untargeted PGD-100 attack, SAWR achieves a robust accuracy of 79.75\% that surpasses OSCP by 9.51\% and maintains a higher clean accuracy (86.10\% vs.\ 81.39\%). This suggests that that our method better preserves essential diagnostic features during the purification process rather than trading clean performance for robustness. 
In the targeted PGD-40 setting, SAWR reaches a robust accuracy of 84.68\%, which outperforms OSCP by 6.03\%. Under the more rigorous AutoAttack evaluation, SAWR improves robust accuracy by 10.69\% (77.77\% vs.\ 67.08\%), which demonstrates that the gains are not specific to gradient-based attacks and generalize to the ensemble threat model.

Across all three attack settings, SAWR achieves consistent gains 
in both robust and clean accuracy over OSCP. This suggests that the improvements are not an artifact of gradient masking but reflect genuine recovery of diagnostic image structure. The simultaneous improvement in clean accuracy indicates that the stain-aware frequency regularization reduces the oversmoothing introduced by the base purification objective and preserves the diagnostic features that the classifier relies upon for both clean and adversarial inputs.



\begin{figure}[tb]
\begin{center}
\subfigure[PGD-100]{\includegraphics[width=0.18\textwidth]{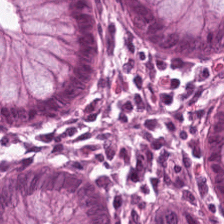}}
  \subfigure[OSCP]{\includegraphics[width=0.18\textwidth]{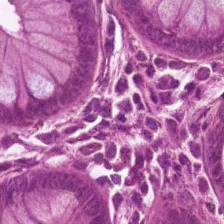}}
  \subfigure[SAWR(Ours)]{\includegraphics[width=0.18\textwidth]{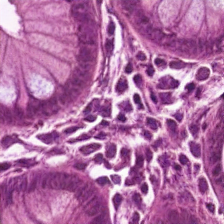}}
  \subfigure[PSD analysis]{\includegraphics[width=0.4\textwidth]{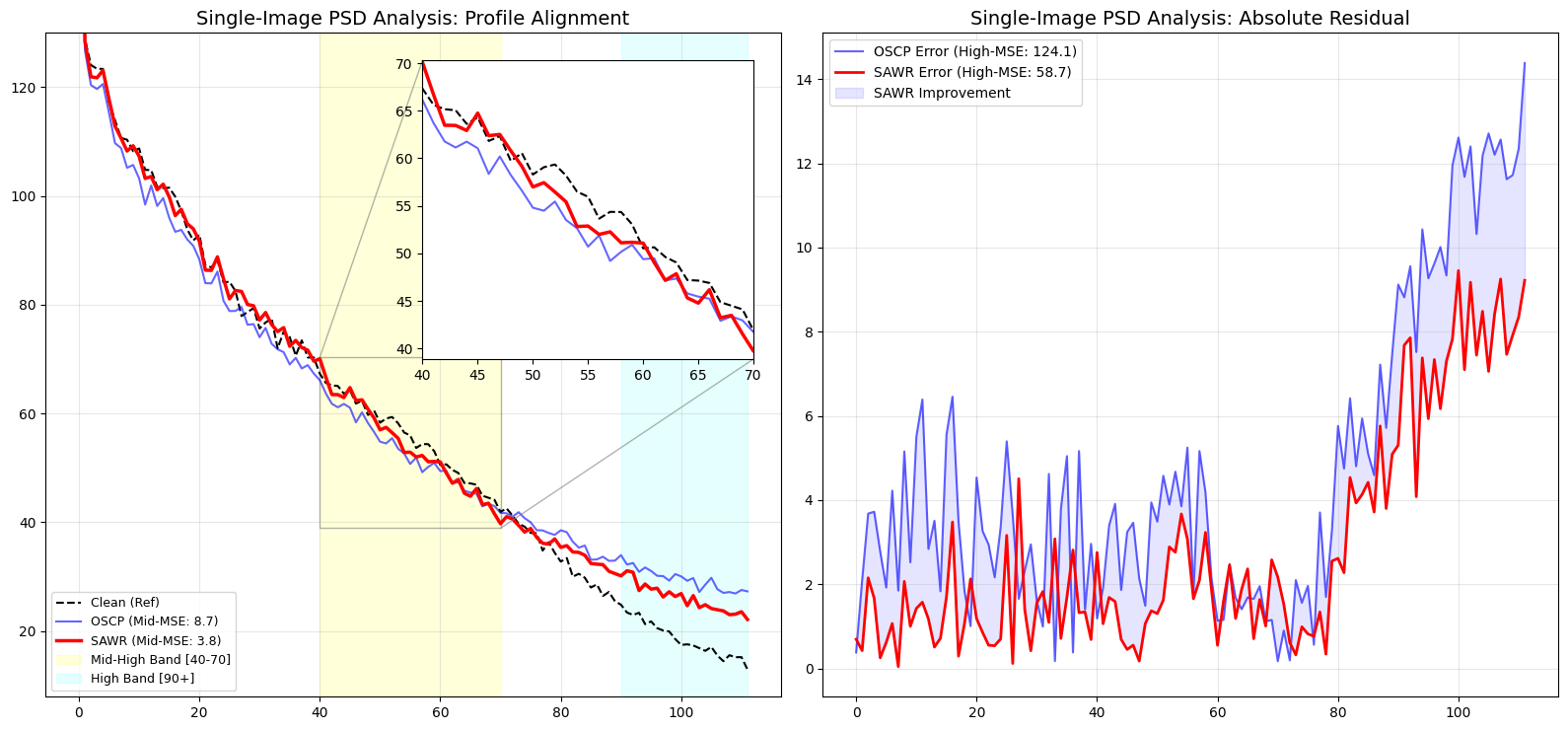}}
  
  \subfigure[PGD-40]{\includegraphics[width=0.18\textwidth]{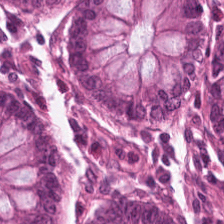}}
  \subfigure[OSCP]{\includegraphics[width=0.18\textwidth]{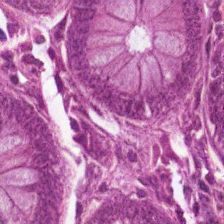}}
  \subfigure[SAWR(Ours)]{\includegraphics[width=0.18\textwidth]{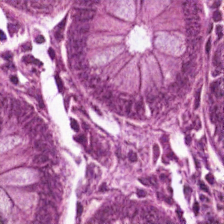}}
  \subfigure[PSD analysis]{\includegraphics[width=0.4\textwidth]{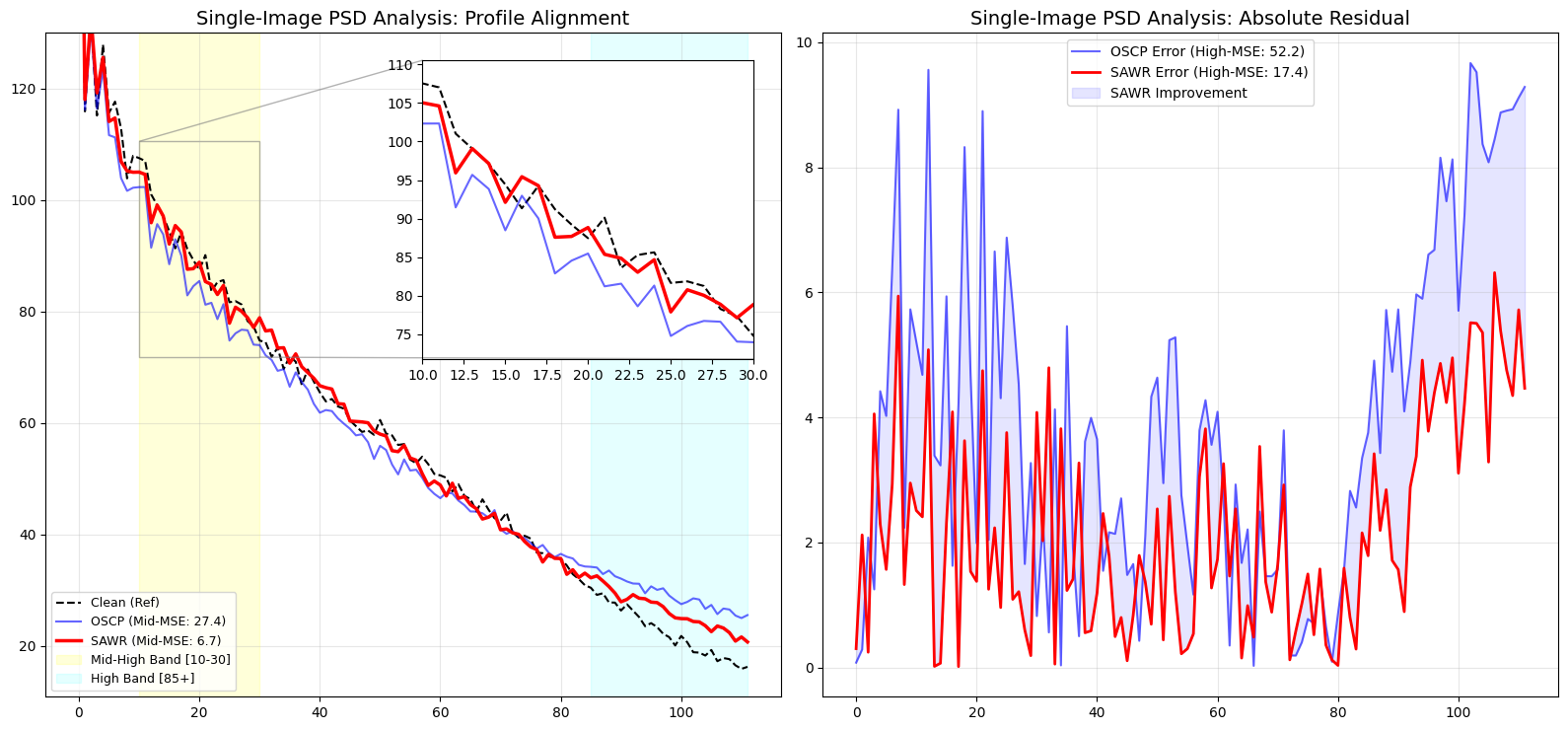}}

  \subfigure[AutoAttack]{\includegraphics[width=0.18\textwidth]{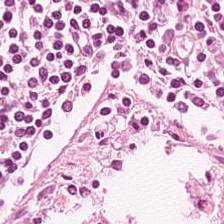}}
  \subfigure[OSCP]{\includegraphics[width=0.18\textwidth]{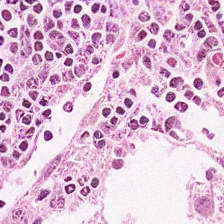}}
  \subfigure[SAWR(Ours)]{\includegraphics[width=0.18\textwidth]{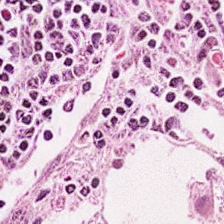}}
  \subfigure[PSD analysis]{\includegraphics[width=0.4\textwidth]{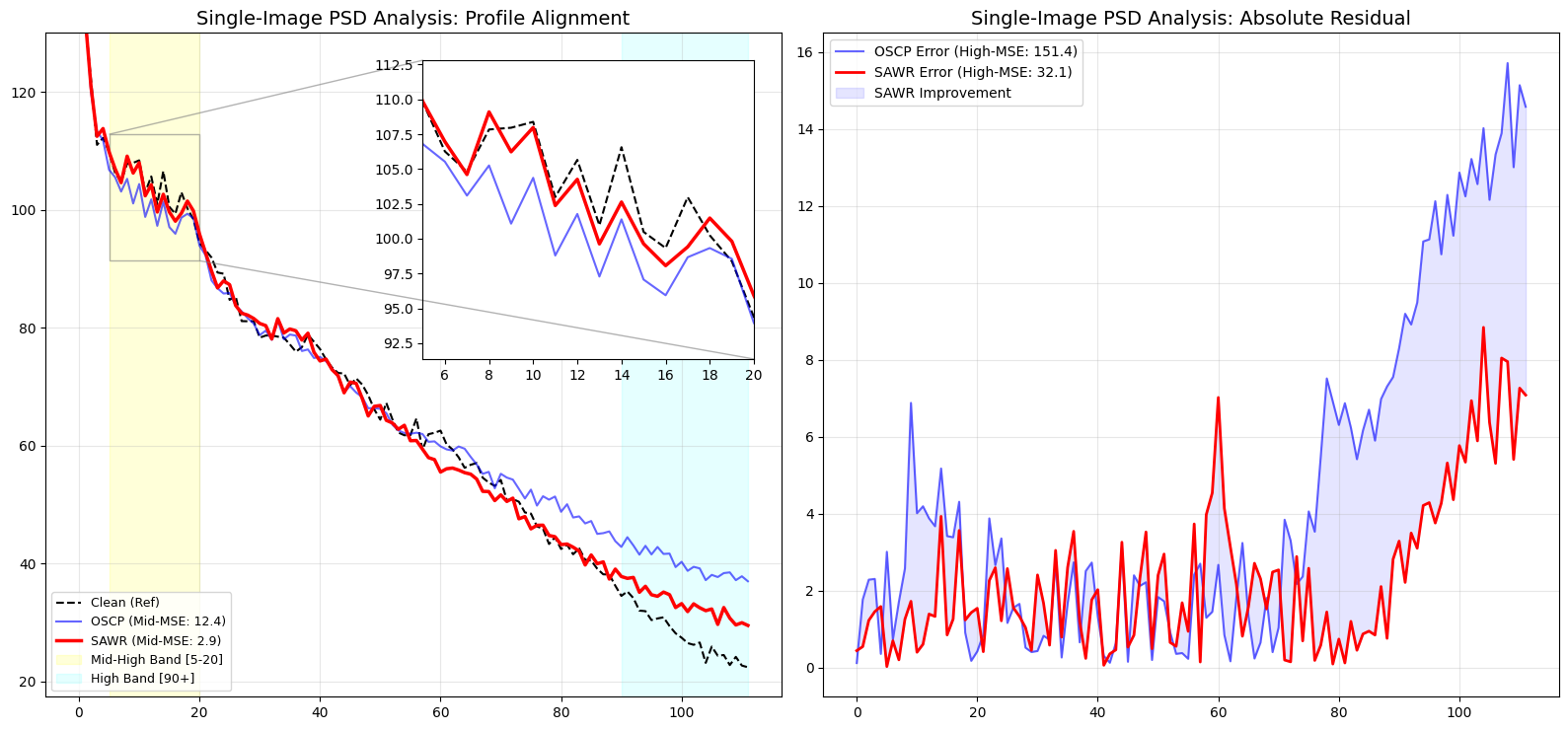}}
\end{center}
\caption{Qualitative comparison and frequency-domain PSD analysis.}
\label{fig:histocomparison}
\end{figure}

\paragraph{Qualitative comparison and frequency analysis.}
To further investigate the purification process, we provide a qualitative comparison and PSD analysis in Fig.~\ref{fig:histocomparison}. 
SAWR enhances nuclear chromatin texture, sharpens glandular boundaries, and reduces the characteristic oversmoothing artifacts observed in the baseline method.
The PSD analysis in Fig.~\ref{fig:histocomparison} (d) provides quantitative support for these observations.
Our SAWR curve (Red) aligns more closely with the clean reference (Black) than the OSCP (Blue) across all frequency bands. As indicated in the residual analysis, SAWR achieves a substantially lower High-Band MSE (17.4 vs. 52.2), confirming that SAWR recovers high-frequency spectral content that OSCP fails to restore.

\begin{figure}[tb]
\begin{center}
\includegraphics[width=\textwidth]{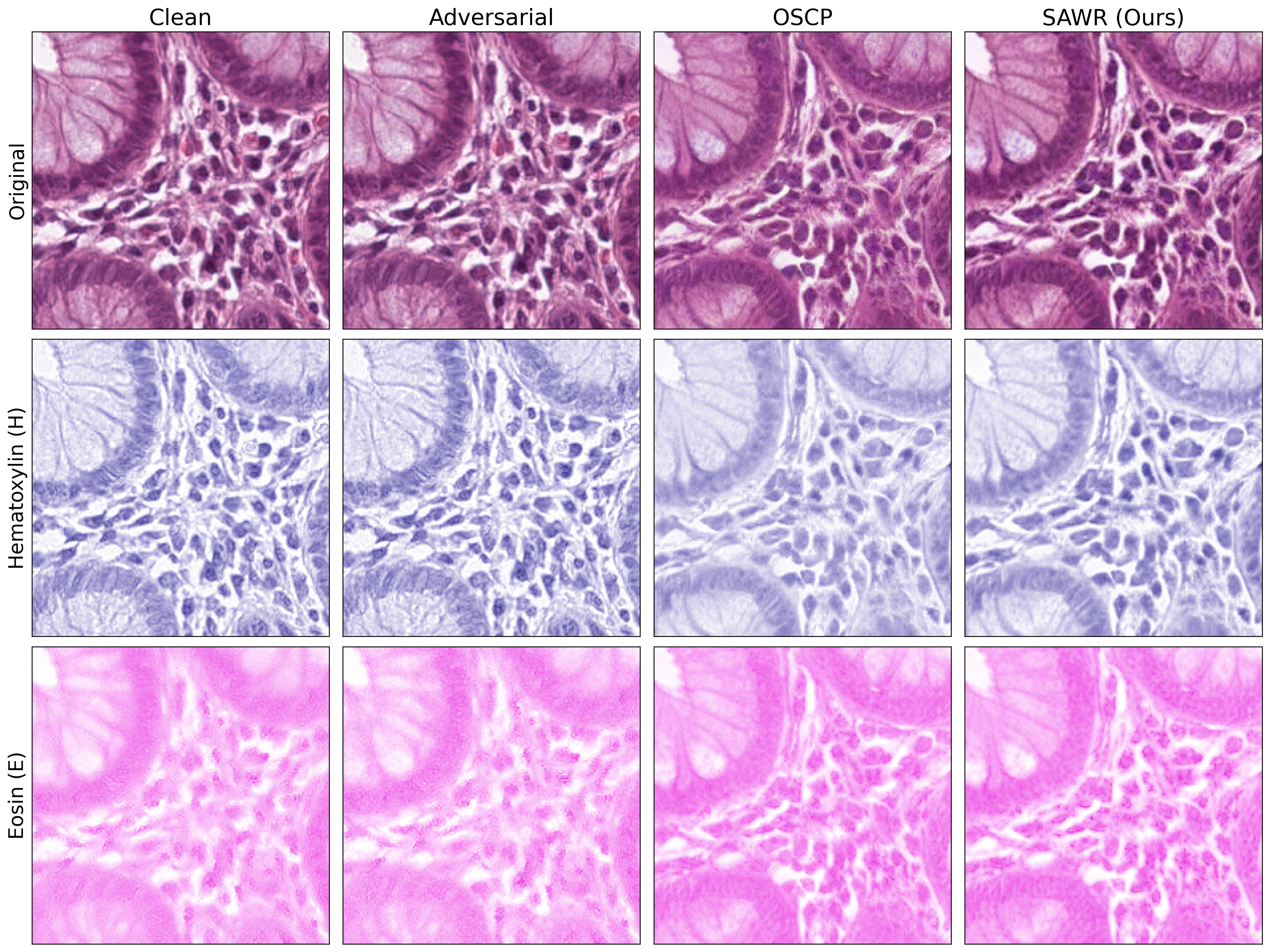}
\end{center}
\caption{Stain decomposition comparison of Clean, Adversarial (PGD-100), OSCP-purified, and SAWR-purified images.}
\label{fig:staindecomposition}
\end{figure}

\paragraph{Stain decomposition analysis.}
Fig.~\ref{fig:staindecomposition} presents stain decomposition 
results obtained via color deconvolution for Clean, Adversarial, 
OSCP-purified, and SAWR-purified images. In the Hematoxylin channel, 
OSCP purification noticeably attenuates nuclear staining intensity 
and reduces chromatin contrast relative to the clean reference. 
SAWR better preserves nuclear staining intensity and chromatin 
boundary detail. This is consistent with the full multi-scale 
wavelet supervision applied to the Hematoxylin channel in 
$\mathcal{L}_{\mathrm{sawr}}$.
In the Eosin channel, the staining distribution remains consistent across all four conditions. This reflects the low-frequency nature of Eosin staining and suggests that global stromal structure is relatively robust to adversarial perturbation at this perturbation budget. The primary challenge in H\&E purification lies in recovering the high-frequency nuclear detail captured by the Hematoxylin channel, where SAWR demonstrates clear improvement over the baseline.


\begin{table}[tb]\setlength{\tabcolsep}{3pt} 
\begin{center}
\resizebox{\columnwidth}{!}{
\begin{tabular}{lcccccccccc}
\toprule
\multirow{2.5}{*}{Method} & \multicolumn{4}{c}{Loss Components} & \multicolumn{2}{c}{Acc (\%)} & \multicolumn{4}{c}{Image Quality} \\
\cmidrule(lr){2-5} \cmidrule(lr){6-7} \cmidrule(lr){8-11}
& $\mathcal{L}_{base}$ & $\mathcal{L}_{mDWT}$ & $\mathcal{L}_{sawr}$  & $\mathcal{L}_{sem}$ & Clean & Rob. & PSNR$\uparrow$ & SSIM$\uparrow$ & Sharp & PSD-MSE$\downarrow$ \\
\midrule
Clean & -- & -- & -- & -- & 94.15 & -- & $\infty$ & 1.000 &  853.3 & 0 \\ 
Adv.  & -- & -- & -- & -- & 94.15 & 0 & 40.64 & 0.9788 &  929.3 & 7.4 \\
OSCP~\cite{lei2025instant} & \checkmark & -- & -- & -- & 81.39 & 70.24 & 20.11 & 0.4666 &  966.6 & 25.56 \\
\midrule
+ $\mathcal{L}_{mDWT}$ & \checkmark & \checkmark & -- & -- & 85.84 & 76.81 & 20.22 & 0.4768 & 907.1 & 19.19 \\
+ $\mathcal{L}_{sawr}$ & \checkmark & \checkmark & \checkmark & -- & 86.48 & 79.11 & 20.22 & 0.4762 & 863.4 & 19.04 \\
\textbf{SAWR} & \checkmark & \checkmark & \checkmark & \checkmark & 86.10 & 79.75 & 20.23 & 0.4775 & 852.6 & 18.62 \\
\bottomrule
\end{tabular}
}
\end{center}
\caption{Ablation study of loss components. The Clean row serves as the statistical target for image fidelity.}
\label{tab:combined_ablation}
\end{table}

\subsection{Ablation study}
\paragraph{Evaluation metrics.}
To evaluate purification quality beyond classification accuracy, we employ four image fidelity metrics: the standard Peak Signal-to-Noise Ratio (PSNR), Structural Similarity Index Measure (SSIM), Sharpness, and Power Spectral Density MSE (PSD-MSE). The latter two are particularly suited to capturing textural and spectral fidelity relative to the clean baseline.
Sharpness is defined by the average gradient magnitude across the image and calculated as the mean response of the Sobel operator, $\nabla I = \sqrt{G_x^2 + G_y^2}$, which reflects the preservation of high-frequency image structures.
A Sharpness value that diverges from the clean reference in either direction indicates either over-smoothing or residual high-frequency noise. 
The PSD-MSE measures the discrepancy between the power spectra of the purified and clean images. It is computed by taking the 2D Discrete Fourier Transform (DFT) of the images \(I(x,y)\), $F(u,v) = \mathcal{F}\{I(x,y)\}$, where \((x,y)\) and \((u,v)\) denote the spatial and frequency coordinates respectively. The mean squared error is then calculated between their radially averaged power profiles $P(f) = |F(u,v)|^2$, which provides a rotation-invariant summary of spectral fidelity.

\paragraph{Loss component analysis.}
To investigate the contribution of each loss component, we evaluate the purified images across four quality metrics as reported in Table~\ref{tab:combined_ablation}.
The OSCP baseline exhibits noticeable degradation in image fidelity, with a PSNR of 20.11, SSIM of 0.4666, Sharpness of 966.6, and PSD-MSE of 25.56. The elevated Sharpness relative to the clean reference (853.3) indicates that OSCP introduces residual high-frequency artifacts rather than restoring the natural spectral profile of clean images.
Adding the multi-level wavelet loss \(\mathcal{L}_{\mathrm{mDWT}}\) without stain adaptation initiates the suppression of these high-frequency artifacts.
The Sharpness metric decreases from 966.6 to 907.1 while PSD-MSE is reduced by 6.37 (from 25.56 to 19.19).
This confirms that explicit frequency-domain supervision provides complementary signal beyond the pixel-level reconstruction objective.

The restoration quality is further improved by incorporating the stain-aware wavelet loss ($\mathcal{L}_{\mathrm{sawr}}$), which adapts the frequency constraints to H\&E stain characteristics. Compared to $\mathcal{L}_{\mathrm{mDWT}}$ alone, $\mathcal{L}_{\mathrm{sawr}}$ brings the Sharpness value substantially closer to the clean reference (863.4 vs.\ 853.3) and achieves a consistent reduction in PSD-MSE (19.04 vs.\ 19.19). This demonstrates that stain-specific frequency decomposition provides more targeted spectral alignment.
Finally, the integration of the semantic feature loss ($\mathcal{L}_{\mathrm{sem}}$) yields the lowest PSD-MSE (18.62) and a Sharpness score (852.6) that closely matches the clean images (853.3). These results indicate that constraining semantic feature similarity has an indirect effect of preserving structural consistency and preventing over-smoothing or texture distortion.

\begin{table}[tb]
\centering
\begin{minipage}[t]{0.38\textwidth}
\centering
\setlength{\tabcolsep}{3pt}
\begin{tabular}{lcc}
\toprule
Method & Clean & Robust \\
\midrule
OSCP & 82.95 & 21.25 \\
SAWR (Ours) & \textbf{85.18} & \textbf{21.91} \\
\bottomrule
\end{tabular}
\caption{Results of adaptive attack under untargeted PGD-100.}
\label{tab:adaptiveattack}
\end{minipage}
\hfill
\begin{minipage}[t]{0.58\textwidth}
\centering
\setlength{\tabcolsep}{3pt}
\begin{tabular}{lcccc}
\toprule
Level & Clean & Robust & PSNR & SSIM \\
\midrule
$J=1$ & 0.8591 & 0.7687 & 20.03 & 0.4635 \\
$J=2$ & \textbf{0.8610} & \textbf{0.7975} & \textbf{20.23} & \textbf{0.4775} \\
\bottomrule
\end{tabular}
\caption{Effect of wavelet decomposition levels on PathMNIST under untargeted PGD-100.}
\label{tab:ablation_levels}
\end{minipage}
\end{table}

\paragraph{Adaptive attack.}
We evaluate robustness under a fully adaptive attack (PGD-100 through the full purification pipeline), with results shown in Table~\ref{tab:adaptiveattack}. SAWR yields a clear improvement in clean accuracy and a slight improvement in robust accuracy, suggesting that the module enhances reconstruction fidelity, which in turn benefits clean accuracy.

\paragraph{Effect of decomposition levels.}
Table~\ref{tab:ablation_levels} compares the performance of different wavelet decomposition level. With single-level $J=1$, the model only supervises the finest-scale detail subbands $\mathcal{D}_1$ and the residual approximation $LL_1$ without further decomposition.
Increasing the decomposition depth to $J=2$ yields consistent improvements across all metrics, with robust accuracy improving from 0.7687 to 0.7975 and SSIM from 0.4635 to 0.4775. These results confirm that the second decomposition level captures frequency content that is both adversarially vulnerable and diagnostically informative, and that single-level supervision is insufficient to recover the mid-frequency spectral fidelity of clean histopathology images.

\paragraph{Sensitivity to stain channel regularization weights.}
Table~\ref{tab:ablationlambdahe} presents the ablation results for the regularization strengths $\lambda_H$ and $\lambda_E$ applied to the stain channels.
Performance remains stable across a reasonable range of these hyperparameters. This indicates that SAWR is not brittle to the exact balance between H and E channel regularization.

\paragraph{Sensitivity to loss weighting hyperparameters}
We conducted a grid search over the $\lambda_{sawr}$ and $\lambda_{sem}$ in main loss, as in table~\ref{tab:ablationweighting}. Both clean accuracy and robust accuracy remain stable across highly diverse weight configurations, which indicates that SAWR does not require heavy dataset specific tuning and is highly resilient to hyperparameter selection.

\begin{table}[tb]
\centering
\begin{minipage}[t]{0.40\textwidth}
\centering
\setlength{\tabcolsep}{2pt}
\small
\begin{tabular}{lcccccc}
\toprule
 & \multicolumn{2}{c}{$\lambda_{E}=0$} & \multicolumn{2}{c}{$\lambda_{E}=1$} & \multicolumn{2}{c}{$\lambda_{E}=2$} \\
\cmidrule(lr){2-3} \cmidrule(lr){4-5} \cmidrule(lr){6-7}
$\lambda_{H}$ & Clean & Rob. & Clean & Rob. & Clean & Rob. \\
\midrule
0 & 83.59 & 78.09 & 83.08 & 77.63 & 83.47 & 78.05 \\
1 & 83.50 & 78.19 & 83.12 & 77.49 & 83.37 & 78.12 \\
2 & 83.41 & 77.99 & \textbf{86.10} & \textbf{79.75} & 84.18 & 79.08 \\
\bottomrule
\end{tabular}
\caption{Stain channel weighting under untargeted PGD-100.}
\label{tab:ablationlambdahe}
\end{minipage}
\hfill
\begin{minipage}[t]{0.56\textwidth}
\centering
\setlength{\tabcolsep}{1.5pt}
\footnotesize
\begin{tabular}{lcccccccc}
\toprule
\multirow{2}{*}{\diagbox[width=3em, height=2.5em]
        {\hspace*{-3pt}\raisebox{-3pt}{$\lambda_{sawr}$}}
        {\hspace*{3pt}\raisebox{3pt}{$\lambda_{sem}$}}} & \multicolumn{2}{c}{0.001} & \multicolumn{2}{c}{0.005} & \multicolumn{2}{c}{0.01} & \multicolumn{2}{c}{0.05}  \\
\cmidrule(lr){2-3} \cmidrule(lr){4-5} \cmidrule(lr){6-7} \cmidrule(lr){8-9}
 & Clean & Rob. & Clean & Rob. & Clean & Rob. & Clean & Rob. \\
\midrule
0.001 & 83.36 & 78.20 & 82.63 & 76.88 & 83.73 & 78.76 & 84.07 & 78.75  \\
0.005 & 83.20 & 77.94 & 82.91 & 77.53 & 83.55 & 78.26 & 84.16 & 78.75 \\
0.01 & 83.33 & 78.06 & 82.81 & 77.47 & \textbf{86.10} & \textbf{79.75} & 83.90 & 78.20 \\
0.05 & 83.57 & 79.05 & 83.87 & 78.94 & 84.25 & 79.46 & 84.43 & 79.40 \\
\bottomrule
\end{tabular}
\caption{Loss weighting under untargeted PGD-100.}
\label{tab:ablationweighting}
\end{minipage}
\end{table}


\section{Conclusion}

In this paper, we present Stain-Aware Wavelet Regularization (SAWR), a novel adversarial purification framework that integrates stain-aware design with multi-level wavelet-based regularization for histopathology image analysis. 
By jointly formulating the purification objective in the wavelet frequency domain and the stain decomposition space, SAWR explicitly decouples high-frequency adversarial noise from mid-frequency diagnostic textures across the Hematoxylin and Eosin channels separately.
The asymmetric frequency supervision strategy reflects the distinct spatial frequency profiles of the two stain components. Full multi-scale regularization is applied to the diagnostically critical Hematoxylin channel, while Eosin regularization is restricted to low-frequency structural consistency, which avoids the over-constraint that would arise from uniform treatment. In addition, a semantic consistency loss in the feature space of a pathology-pretrained extractor enforces purification fidelity simultaneously at the pixel, frequency, and semantic levels.
This multi-level design effectively mitigates the trade-off between robustness and image fidelity and prevents the oversmoothing of fine-grained tissue structures commonly induced by standard reconstruction objectives. 
Extensive experiments on PathMNIST demonstrate that SAWR consistently outperforms the state-of-the-art baseline across untargeted PGD, targeted PGD, and AutoAttack evaluations.
This work highlights the underexplored intersection of adversarial robustness and domain-specific signal structure in medical imaging. By incorporating pathology-specific inductive biases into defense design, SAWR points to a promising direction for future research in robust medical image analysis.

\paragraph{Limitations and future work:} 
A limitation of this study is that our evaluation is restricted to patch-level benchmarks since a first step requires datasets with clear intra-class variation. Extending this validation to multi-institutional WSIs and diverse architectures is a natural next step. Furthermore, a deeper analysis of how the reconstruction fidelity of SAWR contributes to robust accuracy under stronger or more diverse adaptive attack settings motivates an important direction for future work.

\newpage

\paragraph{Acknowledgments:} 
We acknowledge HPC resources from NHR@FAU (projects b143dc, b180dc, y100dc), funded by federal and Bavarian state authorities. NHR@FAU hardware is partially funded by DFG 440719683. Additional support was received from ERC projects MIA-NORMAL 101083647 and CHARMS 101246053, DFG 513220538 and 512819079, and by the Bavarian State Ministry of Science and the Arts (StMWK) through the Bavarian AI Foundation Model Initiative.

\bibliography{egbib}
\end{document}